%% file: main.tex
\documentclass[letterpaper, 10 pt, conference]{ieeeconf}  

\IEEEoverridecommandlockouts 

\usepackage{graphicx}
\usepackage{glossaries}
\usepackage{amsfonts}
\usepackage{caption}
\usepackage{subcaption}
\usepackage{tabularx}
\usepackage{algorithm2e}
\usepackage{textcomp}
\usepackage{tikz}   
\usepackage{algpseudocode}
\RestyleAlgo{ruled}

\usepackage{amsmath}

\newcommand{\mb}[1]{\mathbf{#1}}

\begin{document}

\title{\LARGE \bf Continuum Robot Shape Estimation Using Magnetic Ball Chains}

\author{Giovanni~Pittiglio$^{1,2}$, Abdulhamit~Donder$^2$, and Pierre~E.~Dupont$^2$
\thanks{$^1$ Department of Robotics Engineering, Worcester Polytechnic Insitute (WPI), Worcester, MA 01605, USA. Email: {\tt\small gpittiglio@wpi.edu}
\newline $^2$ Department of Cardiovascular Surgery, Boston Children’s Hospital, Harvard Medical School, Boston, MA 02115, USA. Email:
        {\tt\small \{giovanni.pittiglio, abdulhamit.donder, pierre.dupont\}@childrens.harvard.edu}
        \newline This work was supported by the National Institutes of Health under grant R01HL124020.}%
}

\maketitle

\begin{abstract}
Shape sensing of medical continuum robots is important both for closed-loop control as well as for enabling the clinician to visualize the robot inside the body. There is a need for inexpensive, but accurate shape sensing technologies. This paper proposes the use of magnetic ball chains as a means of generating shape-specific magnetic fields that can be detected by an external array of Hall effect sensors. Such a ball chain, encased in a flexible polymer sleeve, could be inserted inside the lumen of any continuum robot to provide real-time shape feedback. The sleeve could be removed, as needed, during the procedure to enable use of the entire lumen. To investigate this approach, a shape-sensing model for a steerable catheter tip is derived and an observability and sensitivity analysis are presented. Experiments show maximum estimation errors of 7.1\% and mean of 2.9\% of the tip position with respect to total length.
\end{abstract}

\begin{keywords}
Medical Robots and Systems, Steerable Catheters, Flexible Robotics, Magnetic Sensing, Continuum robots.
\end{keywords}

\IEEEpeerreviewmaketitle
\section{Introduction}
\input{sections/introduction}

\section{Shape Estimation Model}
\label{sec:shape_sensing}
\input{sections/shape_sensing}

\section{Observability Analysis}
\label{sec:observability}
\input{sections/observability}

\section{Sensitivity Analysis}
\label{sec:sensitivity}
\input{sections/sensitivity}

\section{Experimental Validation}
\label{sec:experiments}
\input{sections/experiments}

\section{Conclusions}
\label{sec:conclusions}
\input{sections/conclusions}

\bibliographystyle{IEEEtran}
\bibliography{bibliography}											
\end{document}

%% file: sections/introduction.tex
Continuum robots are designed to conform to three-dimensional curves, possessing the ability to alter their form by bending, rotating, and either extending or retracting their structural elements. These flexible capabilities render continuum robots highly suitable for applications that require delicate manipulation and spatial adaptability, such as minimally invasive surgical procedures. While kinematic models have been developed for the various types of continuum robots, effects such as nonlinear elasticity, friction and hysteresis as well as external forces applied by surrounding tissue, degrade the accuracy of these models. 

Shape sensing can provide real-time feedback to augment the kinematic model prediction of robot state and so mitigate modeling inaccuracies. Approaches to real-time sensing that have been studied include fiber Bragg gratings (FBG), electromagnetic (EM) tracking, tendon-based sensing and imaging. FBG sensing is currently receiving much attention due to its ability to provide multiple accurate curvature measurements along a robot’s length, but it remains expensive owing to the cost of the sensors as well as the optical interrogator \cite{Ryu2014, Pittiglio2023d, Pittiglio2023e}. EM sensors can be used effectively in some clinical applications \cite{Ramadani2022}, but are limited to localizing a small number of points along a robot. Inferring shape and external applied forces from tendons and string encoders can provide accurate shape estimates, but requires careful robot design and sensor integration \cite{Orekhov2023}. Shortcomings of image-based shape sensing include cost, resolution limitations and the use of ionizing radiation \cite{Ramadani2022}. 

\begin{figure}
    \centering
    \includegraphics{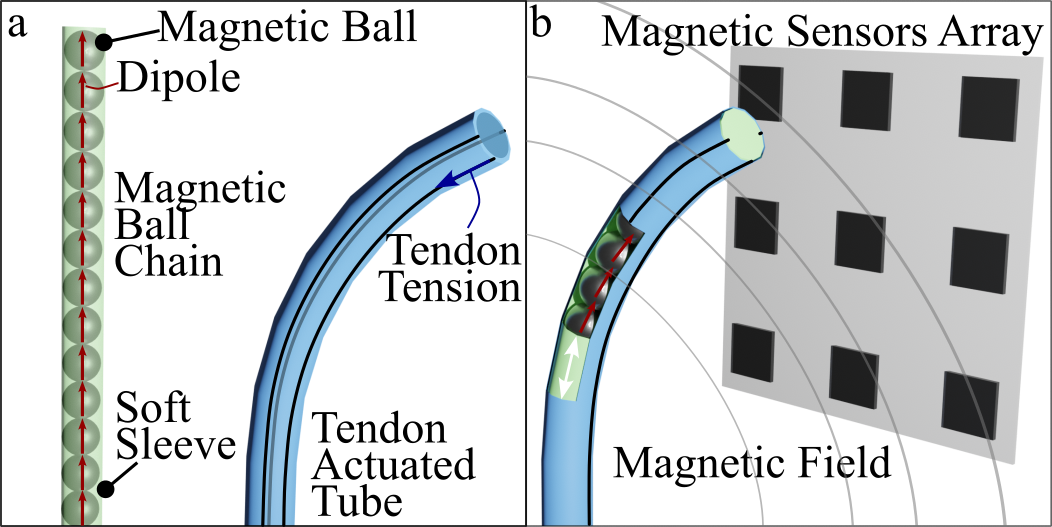}
    \caption{Magnetic ball chain shape sensing. (a) The sensor is comprised of a chain of spherical permanent magnets encased in a flexible sleeve which is inserted in a robot's lumen. (b) An external array of Hall effect sensors measures the combined magnetic field of all balls from which the shape of the robot is computed.}
    \label{fig:platform}
\end{figure}

In contrast to these approaches, a real-time low-cost shape sensing system that could be easily inserted and removed from the robot is preferred because of its direct measurement, instant adaptability to the continuous shape change and ease of replaceability. Magnetic ball chains have recently been introduced as a new design for continuum robots \cite{Pittiglio2023a}. Comprised of a chain of spherical permanent magnets encased in a flexible sleeve, an external magnetic field is used to steer the ball chain. As shown in Fig. \ref{fig:platform}, to create a shape sensor, the ball chain and its sleeve are inserted inside the lumen of a continuum robot and the external magnetic field is replaced by an array of Hall effect sensors. The continuum robot containing the ball chain can be actuated by any of the standard means such as tendons. The ball chain itself is inexpensive, contains no electronics and is easy to sterilize or recycle. 

The potential for tracking permanent magnets embedded in medical robots using external sensor arrays has been established by several researchers. For example, Son \emph{et al.} \cite{Son2019} investigated tracking a capsule robot containing a single permanent magnet. Continuum robot shape estimation has also been performed by attaching a permanent magnet at the tip of each flexing section and interpolating robot shape between magnets \cite{Wang2017, Zhang2017}. Here, we extend the concept by creating a removable chain of spherical magnets which simultaneously maximize the magnetic content and flexibility of the sensor. The goal of our approach is to avoid the need to interpolate robot shape between magnets while also enabling their removal so that the entire robot lumen can be made available for the clinical task after navigation is complete.

In this initial paper, we investigate a simplified shape sensing implementation in which it is assumed that the coordinate frame of the robot base is known and fixed with respect to the world frame. The task is to estimate the shape of a single flexing section that can bend $180^\circ$ in any direction with respect to its longitudinal axis. The remainder of the paper is arranged as follows. The shape sensing model is derived in the next section. The following two sections present observability and sensitivity analyses for inverting the model to estimate shape. Section V contains an experimental evaluation and conclusions appear in the final section.

%% file: sections/shape_sensing.tex
\begin{figure}
    \centering
    \includegraphics[width=\columnwidth]{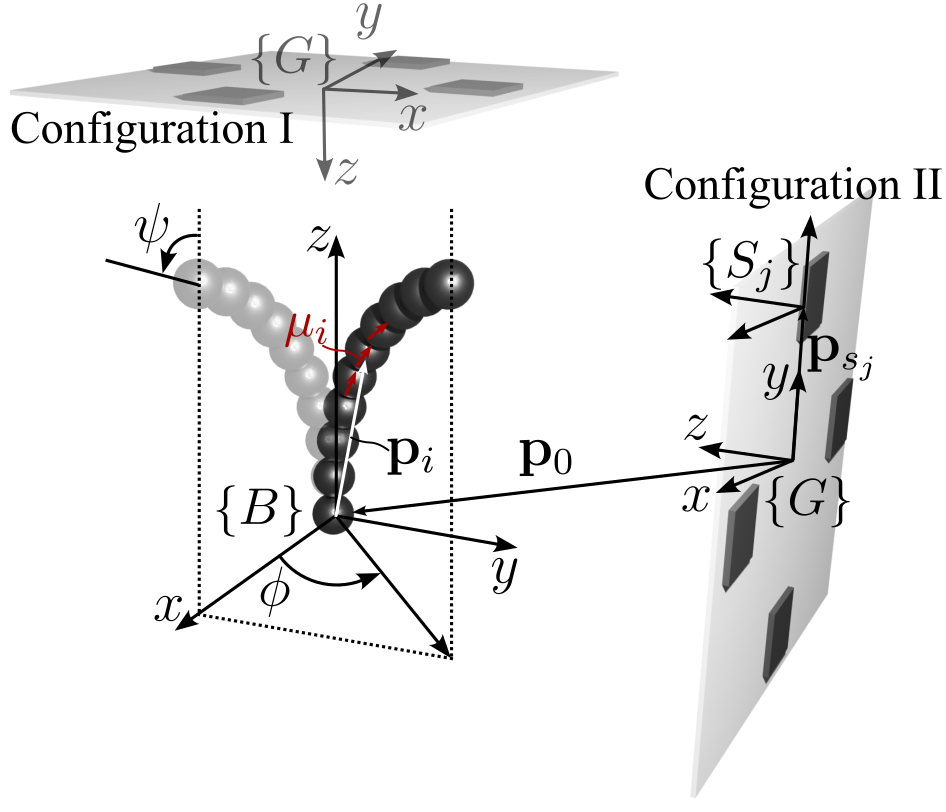}
    \caption{Schematic representation of magnetic ball chain with plane ($\phi$) and angle of bending ($\psi$) parameters. Two possible sensor array locations (I and II) are shown.}
    \label{fig:schematic}
\end{figure}

As shown in Fig. \ref{fig:schematic}, the spheres align to form a chain due to their magnetic interaction \cite{Pittiglio2023a}.  Since the balls are spherical, we can consider each magnet as a perfect dipole \cite{Petruska2013}. When constrained to bend inside a continuum robot, they take on the curved shape of the robot. 

In this paper, we assume that continuum robot flexure is of constant curvature, which we parameterize using angle of bending $\psi$ and plane angle $\phi$ as shown in Fig. \ref{fig:schematic} and assume that each ball's dipole is locally tangent to the curve. Further assuming that the superimposition principle applies, the total field generated can be measured by an array of 3D magnetic field sensors as shown.

We define $\mb q_j \in \mathbb{R}^3$ to be the field measured by the $j$th sensor with respect to the global reference frame $\{G\}$, which equates to the magnetic field generated by the chain at the position of the sensor $\mb p_{s_j}$. We use the magnetic dipole model to compute the field measured by the sensor, for a given bending of the chain 
\begin{equation}
\label{eq:def_conv}
\pmb \gamma = 
\psi \left(
\begin{array}{c}
    -\sin(\phi) \\
    \cos(\phi)
\end{array}
\right).
\end{equation}
The field measured by the $j$th sensor is
\begin{equation}
\label{eq:measure}
    \mb q_j(\pmb \gamma) = {}^W\mb{R}_{S_j} \sum_{i = 1}^n \frac{\mu_0}{4 \pi \|\mb r_{ij}\|^4} \cdot \left(3 \mb r_{ij} \mb r_{ij}^T - I\right){} ^{S_j}\pmb{\mu}_i
\end{equation}
where $n$ is the number of balls in the chain, $\mb p_i$ is the $i$th ball's position, $\pmb{\mu}_i$ its magnetic dipole moment, and we define
\begin{equation}
\mb r_{ij} = \ ^{S_j}\mb p_i + \ ^{S_j}\mb p_0 - \ ^{S_j}\mb p_{s_j}.
\end{equation}
Here $\|\cdot\|$ indicates the Euclidean norm and $\mu_0$ is the vacuum magnetic permeability.

The expression $^F \mb v$ indicates that a vector $\mb v \in \mathbb{R}^3$ is written with respect to the reference frame $F$, and $^{F_1}\mb{R}_{F_2}$ is the rotation from frames $F_2$ to $F_1$.

Under assumption of constant curvature, the position $\mb p_i$ and magnetic dipole $\pmb{\mu}_i$ are
\begin{subequations}
\label{eq:kinematics}
    \begin{equation}
        ^{S_j}\mb p_i ={} ^{S_j}\mb{R}_{G}\left({}^G\mb p_0 +{} ^G\mb R_B\sum_{k = 1}^{i} \exp{\left(\frac{\left[\pmb \gamma\right]^\wedge k}{n}\right) d \mb e_3} \right)
    \end{equation}
    \begin{equation}
        ^{S_j}\pmb \mu_i = {} ^{S_j}\mb R_B\sum_{k = 1}^{i}\exp{\left(\frac{\left[\pmb \gamma\right]^\wedge k}{n}\right) \mu_i \mb e_3} 
    \end{equation}
\end{subequations}
where $d$ is the diameter of the balls, $\mu_i$ the $i$th ball's magnetic dipole intensity, and $\mb e_k$ is the $k$th element of the canonical basis of $\mathbb{R}^3$. The operator $\left[\pmb\gamma\right]^\wedge$ denotes the cross product form of the extended vector $\pmb \gamma' = \left(\pmb \gamma^T \ 0\right)^T$, i.e. $\left[\pmb\gamma\right]^\wedge = \left(\pmb \gamma' \times \mb e_1 \ \pmb \gamma' \times \mb e_2 \ \pmb \gamma' \times \mb e_3 \right)$.

By combining (\ref{eq:measure}) and (\ref{eq:kinematics}), we obtain the total field measured by each sensor. For an array of $m$ sensors, we can combine all measurements as 
\begin{equation}
\mb q(\pmb \gamma) = \left(\mb q_1(\pmb \gamma)^T \ \mb q_2(\pmb \gamma)^T \ \cdots \ \mb q_m(\pmb \gamma)^T\right)^T \in \mathbb{R}^{3m}.
\end{equation}

For a sensor reading $\overline{\mb q}\in \mathbb{R}^{3m}$ we solve for the shape as
\begin{equation}
\label{eq:problem}
    \hat{\pmb \gamma} \ni \mb K(\hat{\pmb \gamma}) \left(\overline{\mb q} - \mb q(\hat{\pmb \gamma})\right) = 0,
\end{equation}
where $\mb K(\hat{\pmb \gamma}) \in \mathbb{R}^{m \times m}$ is a matrix of gains which can be tuned, depending on the confidence level for each sensor. Since different areas of the workspace may have different signal/noise ratio for different sensors, we subdivided the workspace in different manifolds $\Gamma_k$ such that a certain gain $\mb K_k$ is calibrated for $\pmb \gamma \in \Gamma_k$. In our implementation, the manifold $\Gamma_k$ has a center $\overline{\pmb \gamma}_k$ so that 

\begin{equation}
    \pmb \gamma \in \Gamma_k \Leftarrow \min_{k}\| \pmb \gamma - \overline{\pmb \gamma}_k\|.
\end{equation}

We use Algorithm \ref{alg:solution} below to iteratively solve (\ref{eq:problem}) and adjust the sensors gains according to the previous solution. This technique of adjusting the confidence level of each sensor improves the accuracy of the results. Our method for calibrating the  matrices $\mb K_k$ is described in experimental section \ref{sec:experiments}.

\begin{algorithm}
\caption{Shape Estimation Algorithm}\label{alg:solution}
$\mb K(\hat{\pmb \gamma}) \gets$ identity matrix \\
$\hat{\pmb \gamma}^{(0)} \gets$ solution of (\ref{eq:problem}) \\
\While{$l \leq N$}{
  \If{$\pmb \gamma^{(l)} \in \Gamma_k$}{
    $\mb K(\hat{\pmb \gamma}) \gets \mb K_k$ \\
    $\hat{\pmb \gamma}^{(l)} \gets$ solution of (\ref{eq:problem})
  }
}
\KwResult{$\hat{\pmb \gamma}^{(N)}$}
\end{algorithm}

Equation (\ref{eq:problem}) can be solved using gradient-based methods when the Jacobian matrix $\mb J(\pmb \gamma) = \frac{-\partial \mb q(\pmb \gamma)}{\partial \pmb \gamma}$ is full rank. In particular, in the remainder of the paper, we use \emph{lsqnonlin} solver from Matlab (Mathworks, Natick, MA) to solve for $\pmb \gamma$ given a sensor reading of $\overline{\mb q}$. 
Since we assume no prior knowledge of the robot state, we use the rank of the Jacobian to compute the observability of the shape sensing system as described in the next section.

%% file: sections/observability.tex
The bending $\pmb \gamma$ of the robot can be reconstructed from the set of sensor readings $\overline{\mb q}$ as long as the the Jacobian $\mb J(\pmb \gamma)$ is of full rank. Here, we define the reciprocal condition number of $\mb J(\pmb \gamma)$ as 
\begin{equation}
    \chi = \frac{\alpha_m}{\alpha_M},   
\end{equation}
where $\alpha_m$ and $\alpha_M$ are the respective minimum and maximum singular values of $\mb J(\pmb \gamma)$ and use this value to assess the observability of robot curvature. In the plots that follow, $\chi$ is expressed as a percentage.

Since observability will depend on the relative orientation between the robot and the sensor plane, we consider the two scenarios shown in Fig. \ref{fig:schematic}. Configuration I corresponds to the undeflected robot pointing toward the sensor plane. In Configuration II, the undeflected robot points in a direction parallel to the sensor plane. These two configurations can be viewed as a basis set for many clinically relevant robot-sensor configurations.

\begin{figure}
     \centering
     \begin{subfigure}[b]{\columnwidth}
         \centering
         \includegraphics[width=\columnwidth]{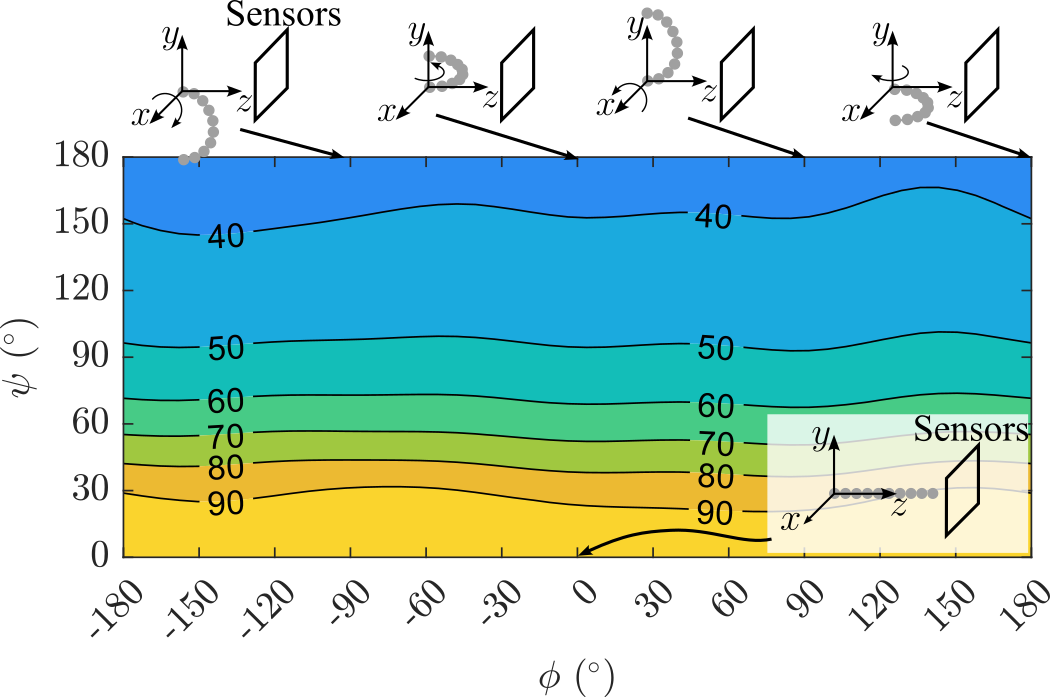}
         \caption{Configuration I}
         \label{subfig:observability_front}
     \end{subfigure}
     \hfill \vspace{0mm}
     \begin{subfigure}[b]{\columnwidth}
         \centering
         \includegraphics[width=\columnwidth]{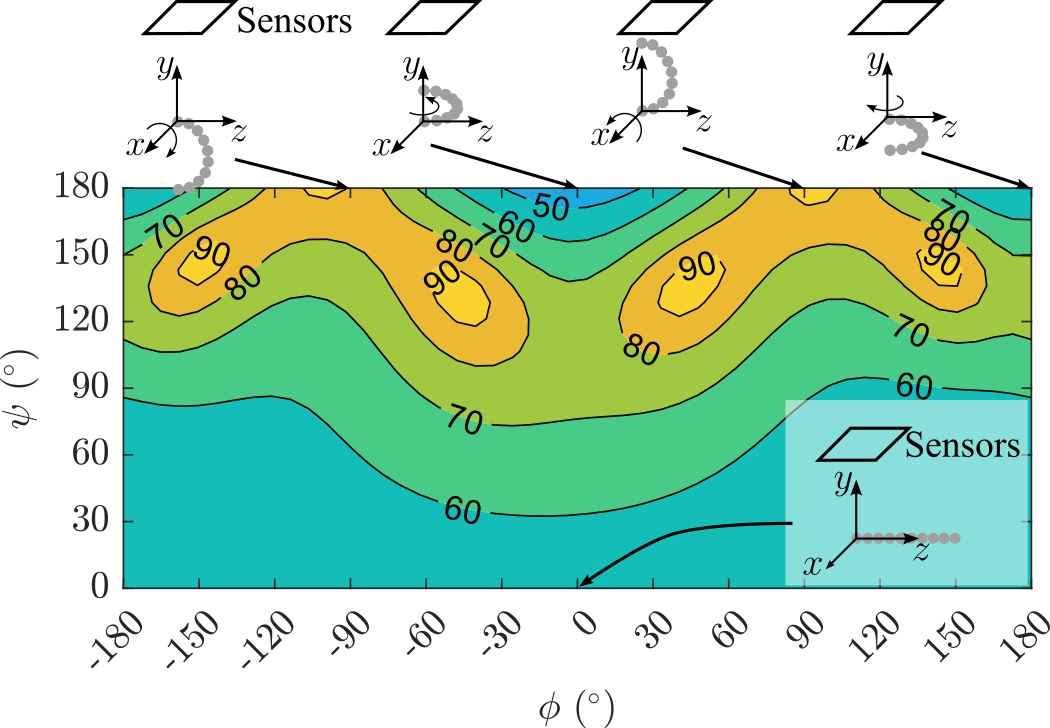}
         \caption{Configuration II}
         \label{subfig:observability_side}
     \end{subfigure}
     \hfill
     
        \caption{Contour plots of shape observability as measured by reciprocal condition number, $\chi$, for the two sensor array configurations shown in Fig. \ref{fig:schematic}. Here, $\chi$ is  expressed as a percentage.}
        \label{fig:observability}
\end{figure}

Figure \ref{fig:observability} presents a numerical assessment of observability for Configurations I and II. In both configurations, the robot base is located at a distance of 15cm from the sensor plane and the robot was allowed to deflect in an arbitrary direction up to $180^\circ$, i.e., $(\phi, \psi) \in \left[-180, 180\right]^\circ \cap \left[0, 180\right]^\circ$. We consider a chain of ten N42 magnetic balls of diameter 6.35mm comparable to those described later in our experiments.

The sensors where positioned at $^G\mb p_{s_j} = 40.5 \cdot \text{rot}_z \left(j\frac{\pi}{2} \right) e_1$ mm, in global frame (see Fig. \ref{fig:schematic}); here $\text{rot}_z\left( \theta \right)$ indicates the rotation of angle $\theta$ around the $z$ axis; $\theta = k \cdot 90^\circ, k = 0, 1, 2, 3$.
 
Figure \ref{subfig:observability_front} plots the reciprocal condition number as function of bending plane and bending angle for Configuration I in which the chain points toward the sensor array when undeflected. Given the symmetry of this configuration with respect to $\phi$, we anticipate that observability will be independent of $\phi$. This can be seen to be true with the highest value of the observability approaching 100\% when the chain is straight and falling off to between 30-40\% when the chain is deflected $180^\circ$. (Note that owing to discretization and numerical rounding, a small amount of variation with respect to $\phi$ is seen in the contours.) This can be anticipated since when the chain is straight, the sensors are well positioned to triangulate on the location of the chain. As it bends, this capability becomes reduced, however, $\chi$ still remains above 30\% indicating that shape estimation is still possible. 

When the sensor array is positioned as shown in Configuration II, the straight configuration is no longer symmetric with respect to all of the sensors resulting in a value of $\chi$ between 50-60\%. In this configuration, however, bending the chain increases observability regardless of the value of $\phi$. 

In particular, the highest values of $\chi$ occur when $\phi \in \{\pm45, \pm90, \pm135\}^\circ$. For $\phi \in \pm90^\circ$, the maxima occur when the robot is deflected $180^\circ$. These configurations correspond to the chain bending directly toward and away from the sensor array. These are again the configurations for which the balls are centrally located with respect to all of the sensors resulting in improved triangulation. 

The maxima at $\phi \in \{\pm45, \pm135\}^\circ$ occur for $\psi \approx 130^\circ$. Given the number of balls and their position and orientation with respect to the sensor array, it is difficult to predict that these are especially observable configurations using intuition alone. The important result is that, while Configuration II would appear to be more challenging for shape estimation than Configuration I, the minimum value of $\chi$ is greater than 40\%. This means that a small change in the chain's configuration is more easily observed when the sensor array is placed to the side of the chain.

%% file: sections/sensitivity.tex
\begin{figure}
    \centering
    \includegraphics[width = 1.0\columnwidth]{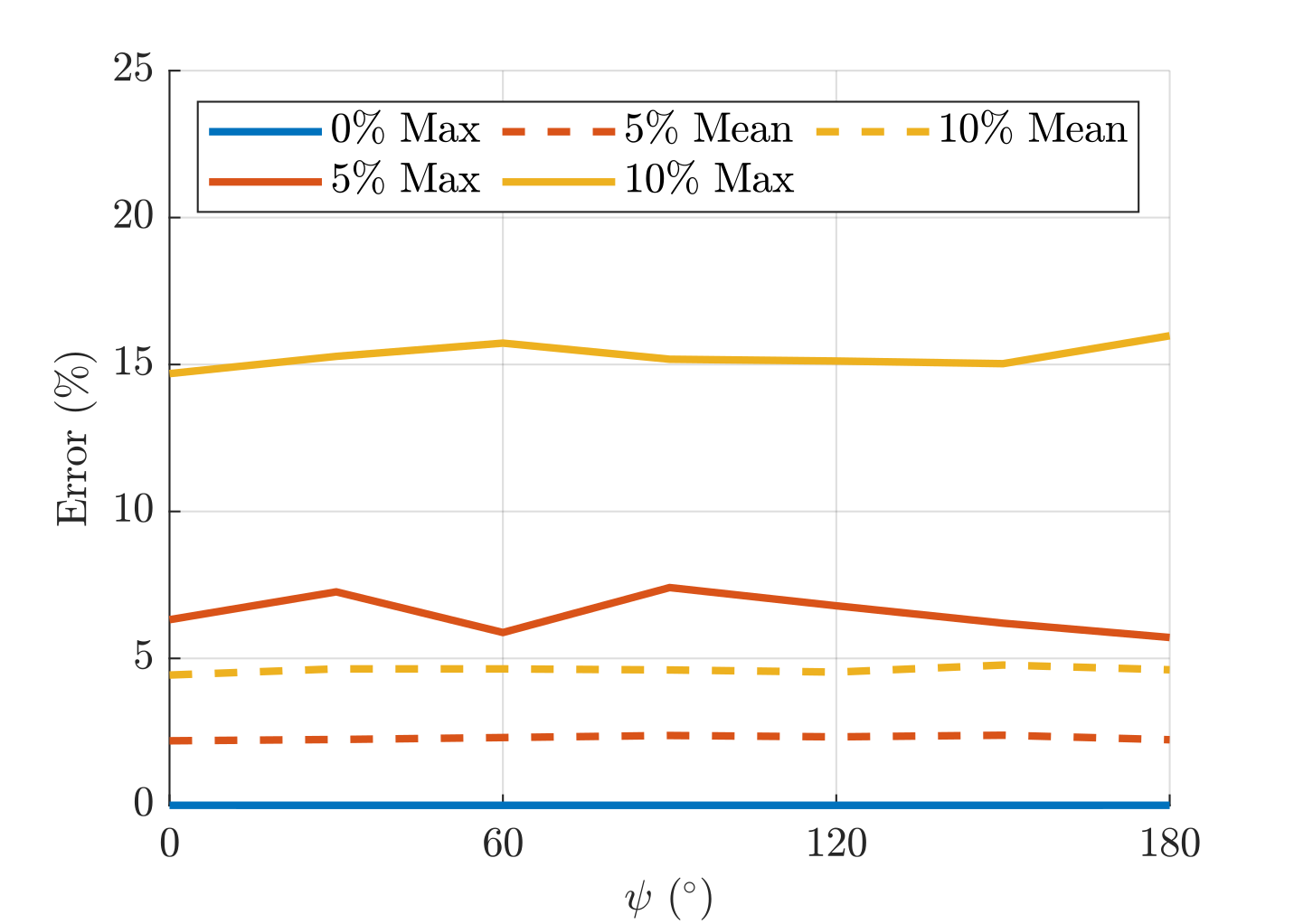}
    \caption{Maximum tip position error (percentage of robot length) as a function of bending angle, $\psi$, for three noise levels in Configuration I (See Fig. \ref{fig:schematic}).}
    \label{fig:sensitivity_front}
\end{figure}

The observability analysis of the preceding section assumes perfect measurements by the Hall effect sensors. In a real sensing system, noise corrupts measurements with small magnitude signals being degraded more than large magnitude signals. To understand how sensor noise would degrade our estimates of robot shape, we performed a sensitivity analysis for Configurations I and II as follows.

We generated simulated measurement data $\overline{\mb q}$ using the model in (\ref{eq:measure}). We discretized the workspace using $(\phi, \psi) = (j\cdot 90^\circ, k\cdot 30^\circ$), $ \forall (j, k) \in [0, 2] \cap [-6, 6]  \cap \mathbb{N}^+$. For each configuration in the workspace, we generated 100 samples, labeled as $l = 1, 2, \dots, 100$, of noisy data by adding zero-mean Gaussian noise with a standard deviation $\sigma = k 5/3 \%$ of the norm of the measured data $\|\overline{\mb q}\|$, with $k = 0, 1, 2$. We used these samples to estimate the maximum measurement errors for 0, 5 and 10\% noise, since 99\% of the samples of the normal distribution are within $3\sigma$.

For the $l$th sample of $k$th chain configuration $\overline{\mb q}_{k_l}$ we solved for the corresponding shape $\hat{\pmb \gamma}_{k_l}$, following Algorithm \ref{alg:solution} for $N = 0$; in this case we have uniform sensors noise and do not need locally tuned gains for the sensors. Using (\ref{eq:kinematics}), we compute the ground-truth position of the chain's tip $\mb p_{k_l} = \mb p_n(\pmb \gamma_{k_l})$ and the estimated tip position $\hat{\mb p}_{k_l} = \mb p_n(\hat{\pmb \gamma}_{k_l})$. We report the error as the maximum norm error over the 100 samples of the tip position 
\begin{equation}
    e_k = \max_{l = 1, 2, \dots, 100}\|\mb p_{k_l} - \hat{\mb p}_{k_l}\|.
\end{equation}

Figure \ref{fig:sensitivity_front} depicts maximum tip position error for Configuration I. 
While, as expected, the error is independent of $\phi$, this plot shows that error is also independent of bending angle $\psi$. While $\chi$ falls as $\psi$ increases, the signal/noise ratio remains high enough that the tip position estimation error is not affected. 

\begin{figure}[t!]
     \centering
     \begin{subfigure}[b]{\columnwidth}
         \centering
         \includegraphics[width=\columnwidth]{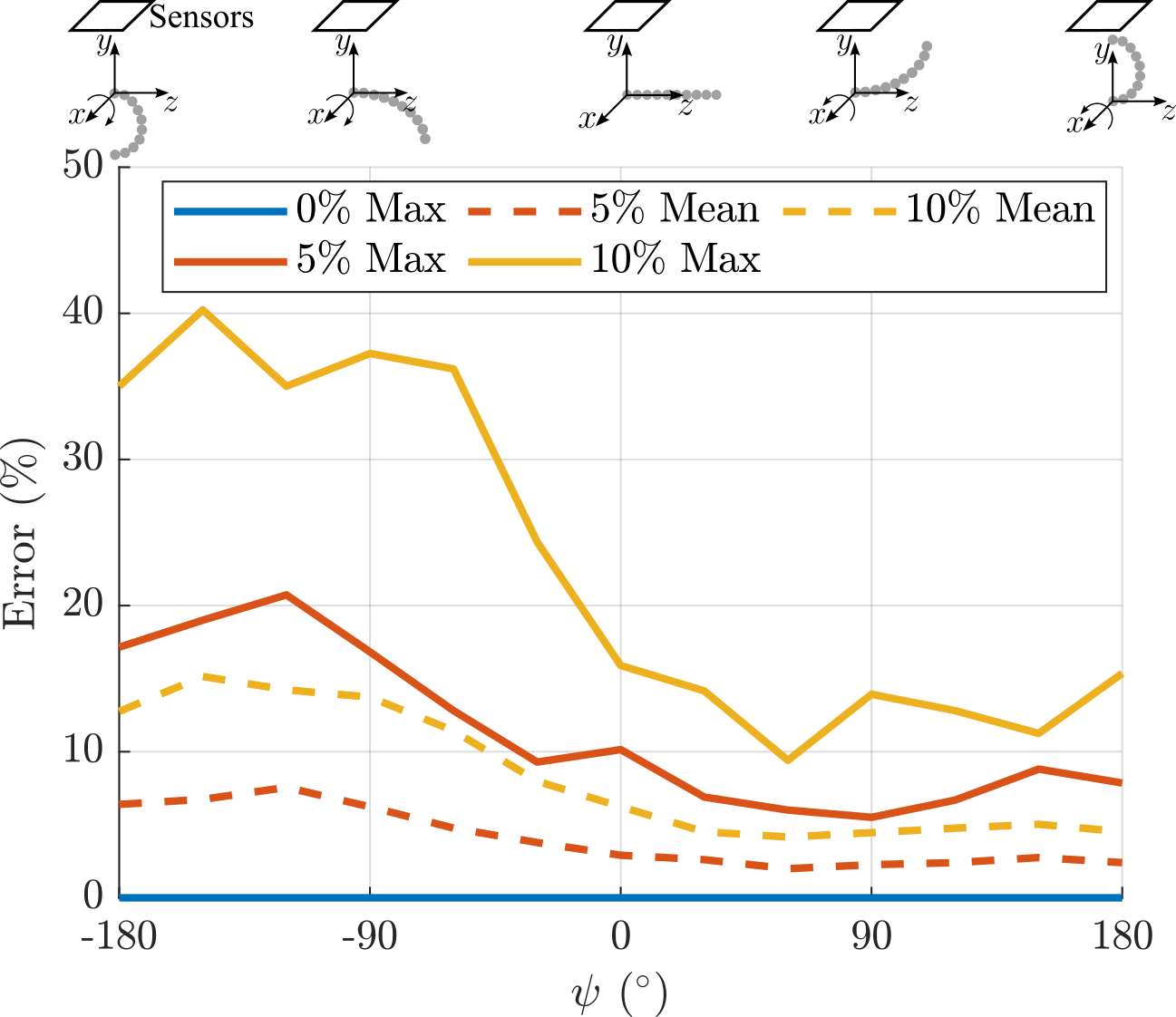}
         \caption{Bending $\pm180^\circ$ in the $y$-$z$ plane (motion toward and away from sensors, $\phi = 0^\circ$).}
         \label{subfig:sensitivity_side_horizontal}
     \end{subfigure}
     \hfill \vspace{0mm}
     \begin{subfigure}[b]{\columnwidth}
         \centering
         \includegraphics[width=\columnwidth]{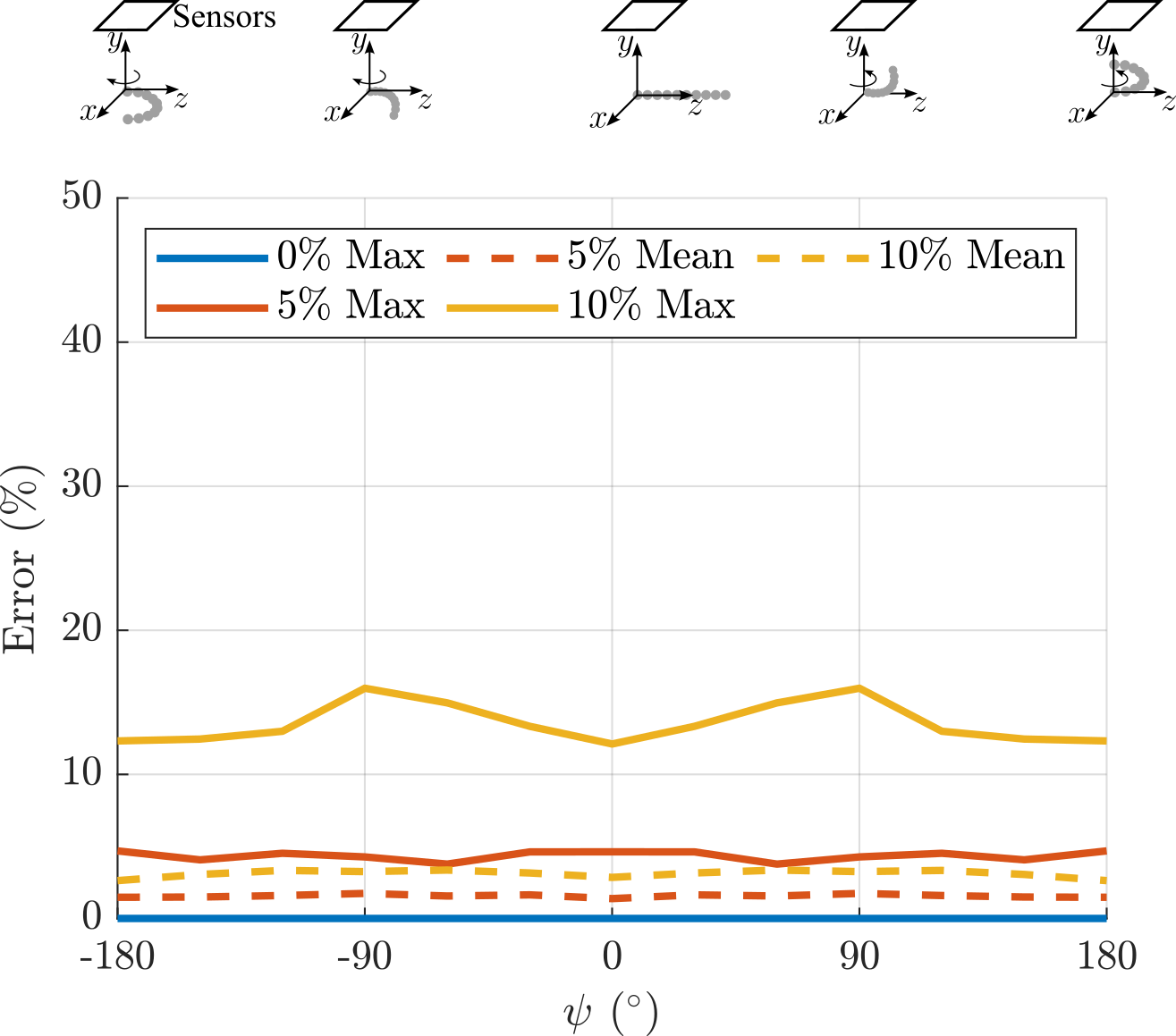}
         \caption{Bending $\pm180^\circ$ in the $x$-$z$ plane (motion parallel to sensors, $\phi = 90^\circ$).}
         \label{subfig:sensitivity_side_vertical}
     \end{subfigure}
     \hfill
        \caption{Maximum tip position error (percentage of robot length) as a function of bending angle, $\psi$, for 3 noise levels in Configuration II (See Fig. \ref{fig:schematic}).}
        \label{fig:sensitivity}
\end{figure}

With the sensor array in Configuration II, maximum tip position error is a function of both bending angle, $\psi$, and robot orientation angle, $\phi$. To illustrate these effects, we plot the results for two planes of bending, $\phi=0^\circ$ and $\phi=90^\circ$, in Fig. \ref{fig:sensitivity}. From Fig. \ref{subfig:sensitivity_side_vertical}, the effect on noise on measurement accuracy can be clearly seen. As the chain bends away from the sensor plane, the signal magnitude falls sufficiently that accuracy degrades significantly. In contrast, Fig. \ref{subfig:sensitivity_side_vertical} shows that robot bending in the plane parallel to the sensor has no effect on estimation accuracy since signal magnitude is comparable over all configurations. 

From this analysis, we can conclude that acceptable maximum estimation errors, within 5\% of chain length, would require sensors with a noise level below 5\% of the signal. Furthermore, as mentioned in Section \ref{sec:shape_sensing}, configuration-dependent weighting of sensor signals can be used to partially compensate for variations in sensor signal amplitude over the workspace.

%% file: sections/experiments.tex
\begin{figure}[t]
    \centering
    \includegraphics[width=1\columnwidth]{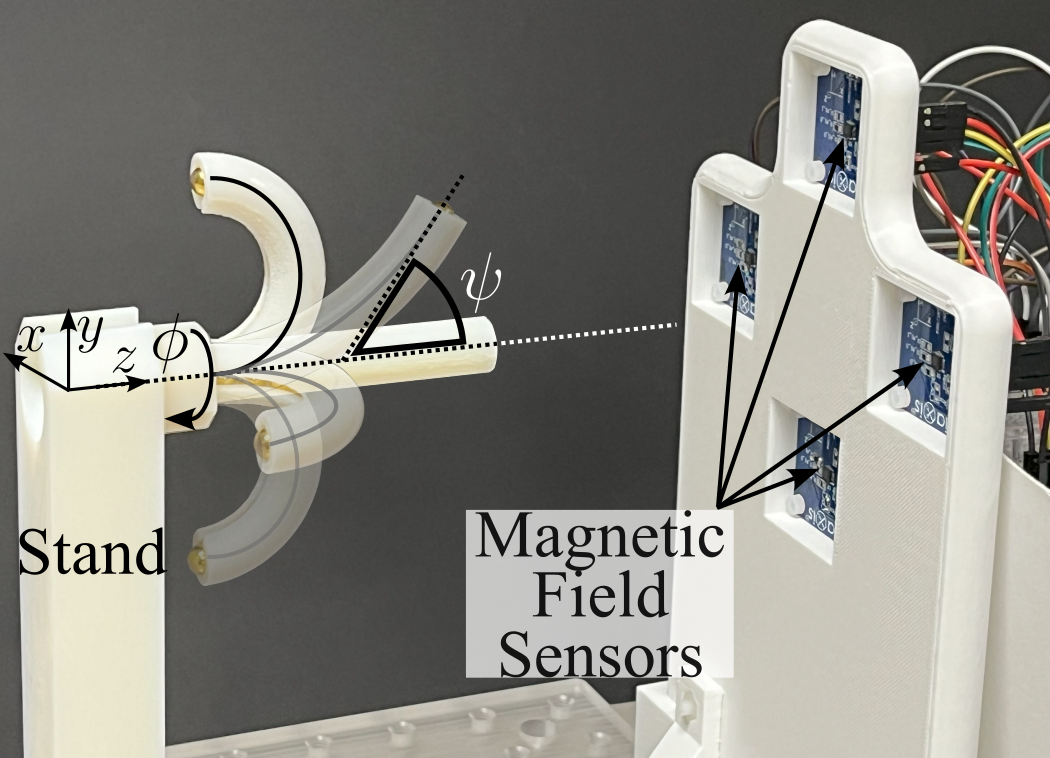}
    \caption{Shape sensing experiments using mock continuum robots of known fixed curvature and orientation.}
    \label{fig:static_setup}
\end{figure}

To evaluate the shape sensing system of Figs. \ref{fig:platform} and \ref{fig:schematic}, we 3D printed continuum ``robots'' of fixed constant curvatures and inserted magnetic ball chains in their lumens as shown in Fig. \ref{fig:static_setup}. This robot orientation matches Configuration I. Since the curvatures of these ``robots'' was known, the values could be used as ground truth and compared to the shape estimated by the sensing system. 

The ball chain consisted of ten N42 spherical magnets of diameter $d = 6.35$mm (K\&J Magnetics), which matched the parameters used in the observability and sensitivity analyses presented above. The sensor array was comprised of four 3D Hall effect senors (MLX90393 Triaxis\textregistered, Melexis), positioned as described in Section \ref{sec:observability}. The mock robot was mounted in a stand which provided a fixed set of possible orientation angles, $\phi$. Matching the analyses above, the stand was positioned such that the proximal ball was 15cm from the sensor plane.

Sensor data was collected for the configurations $(\phi, \psi) = (j\cdot 90^\circ, k\cdot 30^\circ$), $ \forall (j, k) \in [0, 4] \cap [0, 6]  \cap \mathbb{N}^+$. At each configuration, 10 samples for each sensor channel were used for calibration and estimation, as described below. 

\subsection{Sensor Calibration}
Based on the orientation of the chain with respect to the four sensors, the relative accuracy of the sensors varies. To account for this, the optimal gain matrix $\mb K_k$ associated with the manifold of $\Gamma_k$ as described in Section \ref{sec:shape_sensing}, Algorithm \ref{alg:solution}, is computed and used for shape estimation. The center of the manifold $\overline{\pmb \gamma}_k$ is computed from (\ref{eq:def_conv}) for each combination of $(\phi, \psi)_k$. 

\begin{figure}[t!]
     \centering
     \begin{subfigure}[b]{\columnwidth}
         \centering
         \includegraphics[width=\columnwidth]{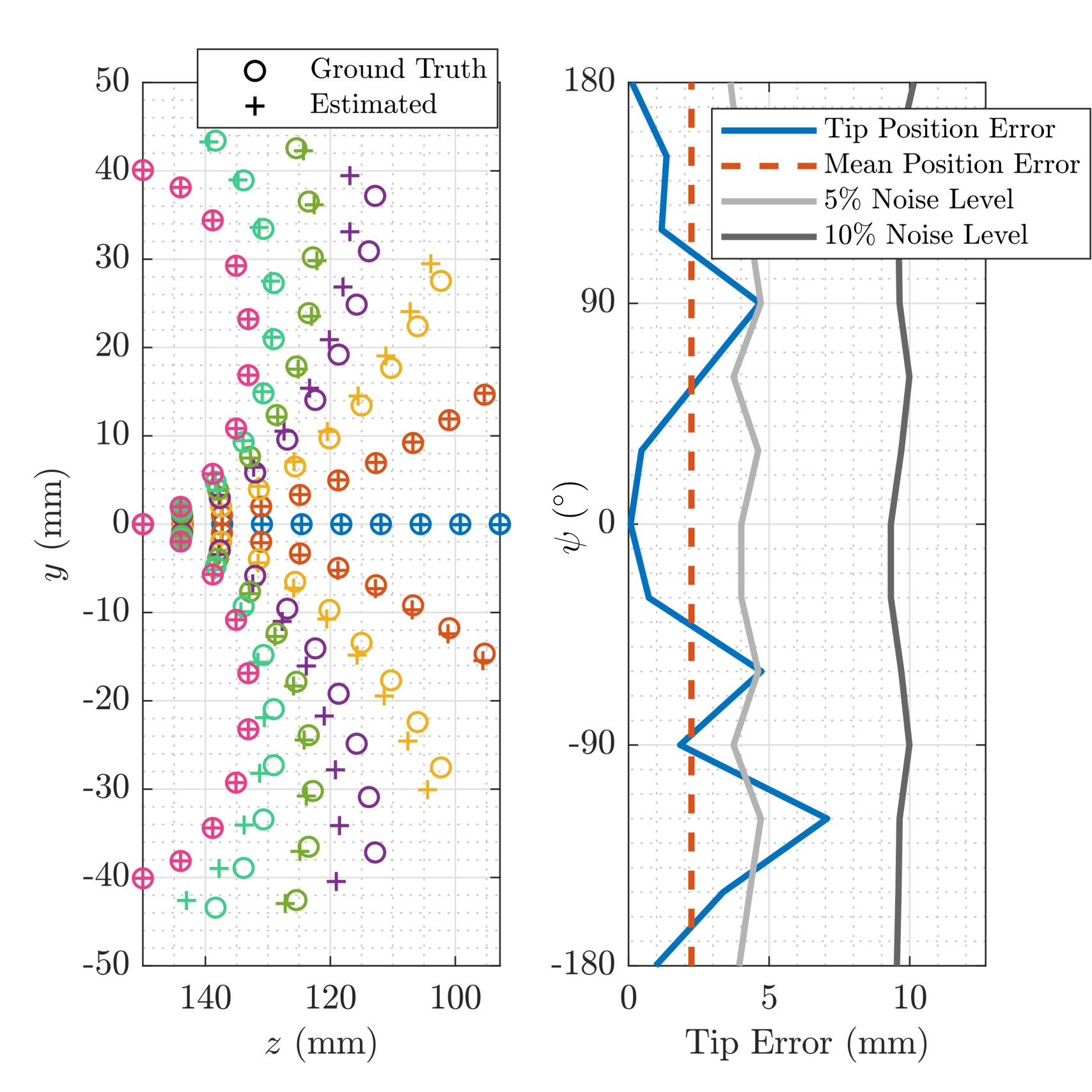}
         \caption{Results on vertical plane: $\phi = 90$.}
         \label{subfig:exp_res_vertical}
     \end{subfigure}
     \hfill \vspace{0mm}
     \begin{subfigure}[b]{\columnwidth}
         \centering
         \includegraphics[width=\columnwidth]{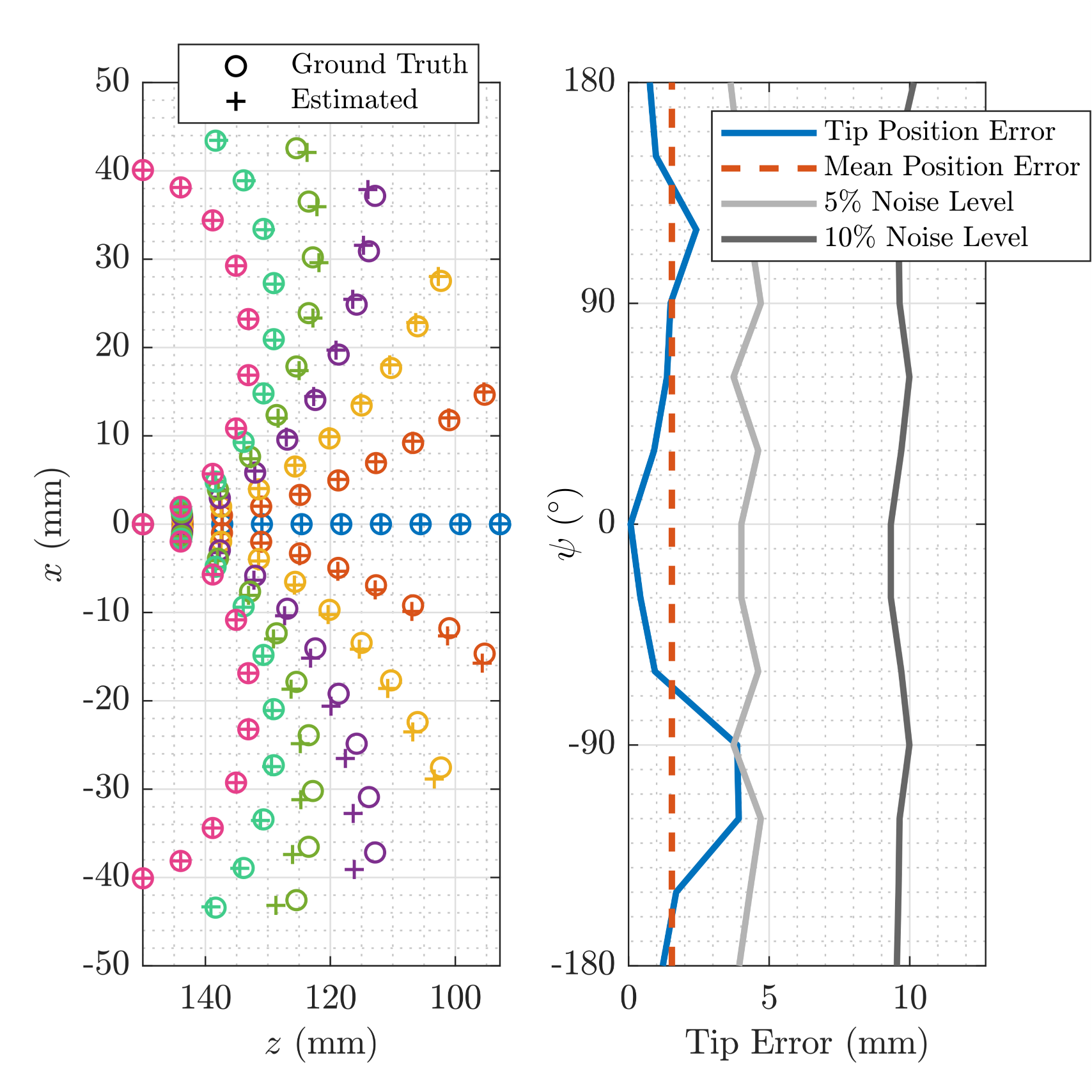}
         \caption{Results on horizontal plane: $\phi = 0^\circ$.}
         \label{subfig:exp_res_horizontal}
     \end{subfigure}
     \hfill
     
        \caption{Experimental estimates of robot shape for Configuration I. Plots on left show estimated and actual ball locations. Plots on right compare sensor-computed tip position errors with predictions from sensitivity analysis of Section IV.}
        \label{fig:exp_res}
\end{figure}
We compute the predicted value of the measurement from the model in (\ref{eq:measure}), $\mb q(\overline{\pmb \gamma}_k)$, and the measurement reading, $\overline{\mb q}_{k_t}$, for each sample $t = 1, 2, \dots, 10$. The maximum error for each component is given by
\begin{equation}
    \mb e_M = \left(\max_{t} |\overline{\mb q}_{k_t} - \mb q(\overline{\pmb \gamma}_k)|\right) \in \mathbb{R}^m.
\end{equation}
where $|\cdot|$ is absolute value and $m = 12$, since four 3D sensors are used.
Sensor gain is defined as
\begin{equation}
    \mb K_k = \left(\text{diag}\left(\mb e_M\right)\right)^{-1} \in \mathbb{R}^{m \times m},
\end{equation}
 and so more trust is given to the channel with smaller maximum error.

\subsection{Shape Sensing}
Algorithm \ref{alg:solution} in Section \ref{sec:shape_sensing} was applied to the collected data with number of iterations $N = 2$. The results are reported in Fig. \ref{fig:exp_res}, where we show independently the bending on the the vertical $y$-$z$ plane ($\phi = 90^\circ$, Fig. \ref{subfig:exp_res_vertical}) and the horizontal $x$-$z$ plane ($\phi = 0^\circ$, Fig. \ref{subfig:exp_res_horizontal}). The plots present the error between the ground truth configurations and the sensor-based estimates. 

For comparison, we also plot the sensitivity analysis maximum error estimates for noise levels of 5\% and 10\% (see Section \ref{sec:sensitivity} for details). Note that the 5\% noise maximum errors bound the actual errors for almost all configurations in both plots. For the vertical plane, the only outlier is the maximum error of 7.1mm for $\psi = -120^\circ$ (11.2\% of the length of the chain). The mean experimental error on the vertical plane is 2.2mm, less than half the size of a magnetic ball, corresponding to 3.5\% of the total length of the chain.

In the horizontal plane of Fig. \ref{subfig:exp_res_horizontal}, the calibration was generally more effective in compensating for sensor noise. All experimental errors fell within the 5\% noise level, with a maximum at $\psi = -90^\circ$ of 3.9mm (6.1\% of the chain length). The mean error on this plane was 1.5mm (2.4\% of the total length).

We also computed an average computation time over all data samples of 0.25s for Algorithm \ref{alg:solution}, indicating that it could be adapted for online use. Alternate real-time techniques such as extended Kalman filters could also be employed to handle dynamic configuration changes.



%% file: sections/conclusions.tex
This paper introduced the concept of using a chain of spherical permanent magnets to generate a shape-specific magnetic field. When inserted in the lumen of a continuum robot, this field can be measured and decoded using an array of Hall effect sensors to estimate the robot shape. Our observability and sensitivity analyses confirm the potential of the method. Our experimental results, using low-cost (\$25/sensor) Hall effect sensors, demonstrated mean tip position errors of 3.5\% of robot length and maximum errors of less than $7.1\%$ of robot length. These results can likely be improved through the use of more accurate sensors. 

While in the case considered here, the base coordinate frame of the robot was assumed known, future work will include estimation of the base frame and also consider robots comprised of multiple curving sections. Furthermore, we will consider optimizing the sensor number and arrangement with respect to the robot in order to optimally balance observability and sensitivity. 

Initial testing not detailed here has shown that the magnetic field is not affected when the chain is inserted inside clinical-grade tendon-actuated catheters. Future work will study catheter compatibility in detail.